\documentclass[11pt]{article}
\pdfoutput=1

\usepackage[preprint]{acl}

\usepackage{times}
\usepackage{latexsym}
\usepackage[T1]{fontenc}
\usepackage[utf8]{inputenc}
\usepackage{microtype}
\usepackage{inconsolata}

\usepackage{graphicx}
\usepackage{booktabs}
\usepackage{multirow}

\usepackage{amsmath}
\usepackage{amsfonts}

\usepackage{enumitem}

\newcommand{\hs}{\textsc{HindSight}}

\title{\hs{}: Evaluating LLM-Generated Research Ideas via Future Impact}

\author{Bo Jiang \\
  Temple University \\
  \texttt{bo.jiang@temple.edu}}

\begin{document}
\maketitle

\begin{abstract}
Evaluating AI-generated research ideas typically relies on LLM judges or human panels---both subjective and disconnected from actual research impact.
We introduce \hs{}, a time-split evaluation framework that measures idea quality by matching generated ideas against real future publications and scoring them by citation impact and venue acceptance.
Using a temporal cutoff~$T$, we restrict an idea generation system to pre-$T$ literature, then evaluate its outputs against papers published in the subsequent 30 months.
Experiments across 10 AI/ML research topics reveal a striking disconnect: LLM-as-Judge finds no significant difference between retrieval-augmented and vanilla idea generation ($p{=}0.584$), while \hs{} shows the retrieval-augmented system produces 2.5$\times$ higher-scoring ideas ($p{<}0.001$).
Moreover, \hs{} scores are \emph{negatively} correlated with LLM-judged novelty ($\rho{=}{-}0.29$, $p{<}0.01$), suggesting that LLMs systematically overvalue novel-sounding ideas that never materialize in real research.
\end{abstract}

\section{Introduction}

Can AI generate research ideas that actually materialize into published work?
Recent systems---ResearchAgent \citep{baek2025researchagent}, AI Scientist \citep{lu2024aiscientist}, and SciMON \citep{wang2024scimon}---demonstrate that large language models can produce coherent research proposals, and a study with 100+ NLP researchers found that LLM-generated ideas are rated as \emph{more novel} than expert-written ones \citep{si2024can}.
But rating an idea as novel is not the same as demonstrating that it anticipates real research.
This gap between \emph{perceived quality} and \emph{actual impact} is the central problem we address.

Current evaluation methods are inherently subjective:
\begin{itemize}[nosep,leftmargin=*]
  \item \textbf{LLM-as-Judge} \citep{zheng2023judging}: scalable but exhibits verbosity bias, self-preference, and poor novelty calibration. Its correlation with real-world impact is unknown.
  \item \textbf{Human evaluation}: expensive, slow, and plagued by low inter-annotator agreement on novelty \citep{si2024can}.
\end{itemize}
Neither approach measures whether a generated idea corresponds to a genuine research direction.

\begin{figure*}[t]
\centering
\includegraphics[width=\textwidth]{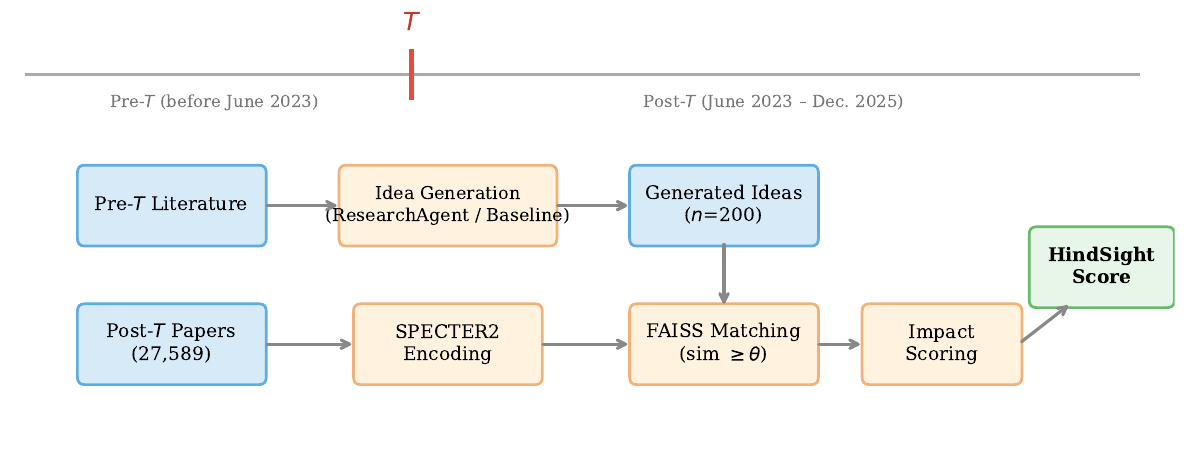}
\caption{The \hs{} framework. An idea generation system accesses only pre-$T$ literature to produce research ideas. These are encoded alongside post-$T$ papers using SPECTER2, matched via FAISS, and scored by the matched papers' real-world citation impact and venue prestige.}
\label{fig:framework}
\end{figure*}

We propose \hs{}, an evaluation framework that provides an \emph{objective}, impact-grounded signal.
The core insight is temporal: by constraining an idea generation system to literature available before a cutoff~$T$, we can evaluate its outputs against papers published \emph{after}~$T$.
Ideas that closely match high-impact future papers score high; those that match nothing score zero (Figure~\ref{fig:framework}).

Applying both \hs{} and LLM-as-Judge to 200 ideas (100 from a retrieval-augmented system, 100 from a vanilla baseline), we find:

\begin{enumerate}[nosep,leftmargin=*]
  \item \textbf{LLM-as-Judge sees no difference}: the two systems receive nearly identical overall scores ($7.44$ vs.\ $7.40$; $p{=}0.584$).
  \item \textbf{\hs{} shows a 2.5$\times$ gap}: the retrieval-augmented system achieves a mean score of $0.297$ vs.\ $0.119$ ($p{<}0.001$).
  \item \textbf{Negative correlation with novelty}: ideas rated as more novel by the LLM are \emph{less} likely to match real future papers ($\rho{=}{-}0.29$, $p{<}0.01$).
\end{enumerate}

\noindent These findings suggest that subjective and objective evaluation capture fundamentally different dimensions of idea quality.
Our contributions are:
\begin{enumerate}[nosep,leftmargin=*]
  \item \hs{}, the first time-split, impact-based evaluation framework for research idea generation (\S\ref{sec:framework}).
  \item Empirical evidence of a systematic disconnect between LLM-as-Judge and objective impact evaluation (\S\ref{sec:results}).
  \item Analysis revealing that LLM judges overvalue ``novel-sounding'' ideas and undervalue ideas that anticipate real research trends (\S\ref{sec:casestudy}).
\end{enumerate}

\section{Related Work}
\label{sec:related}

\paragraph{Research Idea Generation.}
ResearchAgent \citep{baek2025researchagent} iteratively retrieves scientific literature to produce research proposals using an LLM backbone for problem identification and method development.
The AI Scientist \citep{lu2024aiscientist} extends this to a full pipeline---implementing ideas as code and generating complete papers.
SciMON \citep{wang2024scimon} optimizes for novelty by contrasting generated ideas against existing work.
\citet{si2024can} conducted a large-scale study finding that LLM-generated ideas are rated as more novel but less feasible than expert ideas.
These systems demonstrate increasing sophistication in generation, but evaluation remains the bottleneck.

\paragraph{Evaluation Methods.}
LLM-as-Judge \citep{zheng2023judging} is the dominant paradigm, with LLMs rating ideas on dimensions like novelty and impact.
Known limitations include verbosity bias, self-enhancement bias, and poor calibration on open-ended judgments.
Human evaluation provides a complementary signal but shows high variance---\citet{si2024can} report particularly low inter-annotator agreement on novelty.
\citet{li2024automated} find significant discrepancies between LLM and human paper reviews.
To our knowledge, no prior work has proposed an evaluation framework grounded in \emph{real-world research outcomes}.

\paragraph{Time-Split Evaluation.}
Evaluating predictions against future outcomes is standard in finance (backtesting) and recommender systems (temporal splits).
In NLP, temporal splits prevent data contamination in knowledge-intensive tasks.
\hs{} applies this principle to research ideas: we treat post-$T$ publications as ground truth for assessing ideas generated with pre-$T$ knowledge.

\section{The \hs{} Framework}
\label{sec:framework}

\subsection{Problem Formulation}

Let $G$ be an idea generation system with access to literature $\mathcal{L}_{<T}$ published before time~$T$, producing ideas $\mathcal{I} = G(\mathcal{L}_{<T})$.
Each idea $i \in \mathcal{I}$ consists of a problem statement and a proposed method.
Let $\mathcal{P}_{>T}$ be the \emph{ground truth pool} of papers published after~$T$.
The goal is to evaluate how well ideas in $\mathcal{I}$ anticipate research in $\mathcal{P}_{>T}$.

\subsection{Time-Split Design}
\label{sec:timesplit}

The cutoff $T$ must satisfy two constraints:
(1)~the LLM's knowledge cutoff falls \emph{after}~$T$ with a safety margin to prevent information leakage, and
(2)~the ground truth window $[T, T{+}\Delta]$ is long enough to capture meaningful developments.
We use $T{=}\text{June 2023}$ with Llama-3.3-70B-Instruct (cutoff: December 2023), giving a 6-month margin and a 30-month ground truth window.

\subsection{Matching}
\label{sec:matching}

For each idea $i$, we identify matching papers via semantic similarity:

\begin{enumerate}[nosep,leftmargin=*]
  \item \textbf{Encode}: represent ideas as \emph{problem} $\oplus$ \emph{method} and papers as \emph{title} $\oplus$ \emph{abstract}, where $\oplus$ is concatenation with a separator.
  \item \textbf{Retrieve}: find the top-$K$ most similar papers from $\mathcal{P}_{>T}$.
  \item \textbf{Filter}: retain papers with cosine similarity ${\geq}\,\theta$, forming the match set:
\end{enumerate}
\vspace{-2mm}
\begin{equation}
  \mathcal{M}(i) = \{ p \in \text{top-}K(\mathcal{P}_{>T}, i) \mid \text{sim}(i, p) \geq \theta \}
\end{equation}

\subsection{Impact Scoring}

Each paper $p \in \mathcal{P}_{>T}$ receives an impact score:
\begin{equation}
  h(p) = 0.6 \cdot \hat{c}(p) + 0.4 \cdot v(p)
\end{equation}
where $\hat{c}(p)$ is the min-max normalized citation count within $\mathcal{P}_{>T}$, and $v(p) \in \{0, 1\}$ indicates publication at a top venue (ICLR, NeurIPS, ICML, ACL, EMNLP, CVPR, or AAAI).

The \hs{} score is the maximum impact among matched papers:
\begin{equation}
  \hs{}(i) = \max_{p \in \mathcal{M}(i)} h(p)
\label{eq:hindsight}
\end{equation}
If $\mathcal{M}(i) = \emptyset$, then $\hs{}(i) = 0$.
We use the maximum rather than average because a single high-impact match provides strong evidence that the idea anticipated a significant direction.

\section{Experimental Setup}
\label{sec:setup}

\subsection{Ground Truth Pool}

We query the Semantic Scholar API \citep{kinney2023semantic} for AI/ML papers published between June 2023 and December 2025 across 10 research topics (Appendix~\ref{sec:topics}).
After deduplication, the pool contains \textbf{27,589 unique papers} with titles, abstracts, citation counts, and venue information.

\subsection{Idea Generation Systems}

\paragraph{ResearchAgent (retrieval-augmented).}
We implement a simplified two-stage version of ResearchAgent \citep{baek2025researchagent}: a \emph{ProblemIdentifier} reads a seed paper and surfaces open problems, and a \emph{MethodDeveloper} proposes a concrete approach.
Both stages retrieve additional pre-$T$ papers via the Semantic Scholar API (restricted to before June 2023).
We select 10 seed papers per topic and generate one idea per seed, yielding \textbf{100 ideas}.

\paragraph{Vanilla baseline (no retrieval).}
The same LLM is prompted with the topic name and a generic instruction to propose a research idea, producing \textbf{100 ideas} (10 per topic) without literature retrieval.

Both systems use Llama-3.3-70B-Instruct \citep{grattafiori2024llama} served via vLLM \citep{kwiatkowski2023vllm}.

\subsection{Embedding and Matching}

We encode all documents using SPECTER2 \citep{cohan2020specter,singh2023scirepeval}, a transformer pre-trained on citation graphs (768-dim CLS embeddings).
Matching uses a FAISS \citep{johnson2021billion} inner-product index over L2-normalized vectors (cosine similarity), retrieving top-$K{=}20$ per idea.

\paragraph{Threshold calibration.}
SPECTER2 base produces highly concentrated similarity distributions for AI/ML text (0.91--0.98 range).
We select $\theta{=}0.96$ through sensitivity analysis (\S\ref{sec:threshold}), which provides strong discrimination while retaining enough matches for meaningful analysis.

\subsection{LLM-as-Judge}

All 200 ideas are scored by Qwen3-32B \citep{yang2025qwen3} on four dimensions (1--10): Novelty, Feasibility, Expected Impact, and Overall quality.
Each idea is evaluated 3 times ($T{=}0.7$) and scores averaged.
We deliberately use a different model family from the generator to avoid self-preference bias.

\section{Results}
\label{sec:results}

\subsection{Main Results}

\begin{figure*}[t]
\centering
\includegraphics[width=\textwidth]{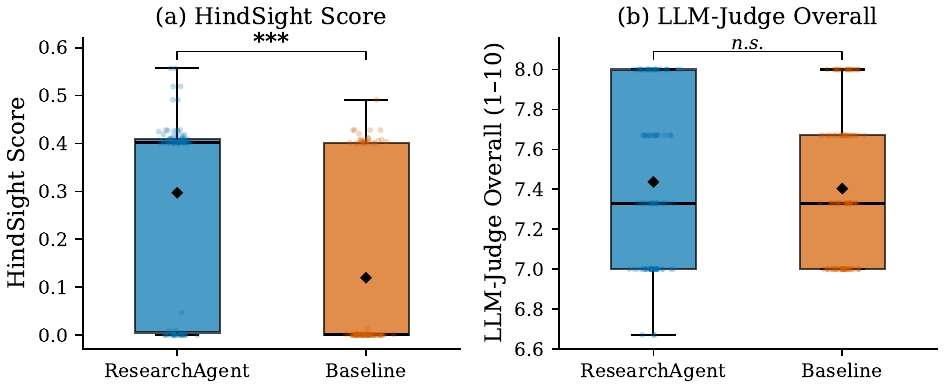}
\caption{Score distributions for both evaluation methods. \textbf{(a)}~\hs{} clearly separates the two systems, with the baseline clustering at zero. \textbf{(b)}~LLM-as-Judge Overall scores are nearly identical ($p{=}0.584$). Diamond markers show means.}
\label{fig:comparison}
\end{figure*}

Table~\ref{tab:main} and Figure~\ref{fig:comparison} present the comparison between the retrieval-augmented system (RA) and vanilla baseline (BL).

\begin{table}[t]
\centering
\small
\begin{tabular}{lcccc}
\toprule
\textbf{Metric} & \textbf{RA} & \textbf{BL} & $\boldsymbol{\Delta}$ & $\boldsymbol{p}$ \\
\midrule
\multicolumn{5}{l}{\emph{\hs{} evaluation}} \\
\quad Score (mean) & \textbf{0.297} & 0.119 & +0.178 & {$<$0.001} \\
\quad Score (median) & \textbf{0.403} & 0.000 & +0.403 & --- \\
\quad Match rate & \textbf{81\%} & 42\% & +39\% & --- \\
\quad Avg.\ matches & \textbf{9.0} & 3.3 & +5.7 & --- \\
\midrule
\multicolumn{5}{l}{\emph{LLM-as-Judge evaluation}} \\
\quad Overall & 7.44 & 7.40 & +0.03 & 0.584 \\
\quad Novelty & 6.68 & \textbf{7.11} & $-$0.42 & {$<$0.001} \\
\quad Impact & 7.87 & \textbf{8.17} & $-$0.30 & {$<$0.001} \\
\quad Feasibility & \textbf{7.68} & 7.09 & +0.59 & --- \\
\bottomrule
\end{tabular}
\caption{System comparison (RA = ResearchAgent, BL = baseline). Statistical tests: Mann--Whitney $U$. \hs{} sharply distinguishes the two systems, while LLM-as-Judge Overall does not.}
\label{tab:main}
\end{table}

The disconnect is stark.
\hs{} reveals that 81\% of retrieval-augmented ideas match at least one ground truth paper, compared to only 42\% for the baseline, with a 2.5$\times$ higher mean score ($p{<}0.001$).
LLM-as-Judge, in contrast, finds \emph{no significant difference} in overall quality ($p{=}0.584$).

The per-dimension LLM scores are equally revealing: the baseline is rated \emph{higher} on both novelty ($p{<}0.001$) and expected impact ($p{<}0.001$).
The only dimension favoring the retrieval-augmented system is feasibility---consistent with its ideas being more grounded in specific literature.

\subsection{Correlation Analysis}

Table~\ref{tab:correlation} reports Spearman correlations \citep{spearman1904proof} between \hs{} scores and each LLM-Judge dimension.

\begin{table}[t]
\centering
\small
\begin{tabular}{lcccc}
\toprule
& \multicolumn{2}{c}{\textbf{ResearchAgent}} & \multicolumn{2}{c}{\textbf{Baseline}} \\
\cmidrule(lr){2-3} \cmidrule(lr){4-5}
\textbf{Dimension} & $\rho$ & $p$ & $\rho$ & $p$ \\
\midrule
Novelty     & $-$0.291 & 0.003 & $-$0.140 & 0.164 \\
Feasibility & +0.252   & 0.012 & +0.006   & 0.951 \\
Impact      & $-$0.225 & 0.025 & $-$0.150 & 0.135 \\
Overall     & $-$0.075 & 0.457 & $-$0.124 & 0.219 \\
\bottomrule
\end{tabular}
\caption{Spearman $\rho$ between \hs{} and LLM-Judge dimensions. For ResearchAgent, \hs{} is \emph{negatively} correlated with novelty and impact but positively correlated with feasibility. All baseline correlations are non-significant.}
\label{tab:correlation}
\end{table}

The most striking pattern is the \textbf{negative correlation with novelty} ($\rho{=}{-}0.29$, $p{=}0.003$): ideas the LLM rates as more original are \emph{less} likely to match real future work.
This suggests LLMs confuse surface-level originality (``novel-sounding'') with genuinely anticipatory thinking (aligned with actual research trajectories).

The positive correlation with feasibility ($\rho{=}+0.25$, $p{=}0.012$) is intuitive: practically grounded ideas more closely resemble real research.
The negative correlation with LLM-judged impact ($\rho{=}{-}0.22$, $p{=}0.025$) further underscores the disconnect---perceived ambition and actual impact point in opposite directions.

For the baseline, all correlations are non-significant, likely reflecting a more homogeneous set of ideas with less quality variation.

\subsection{Threshold Sensitivity}
\label{sec:threshold}

\begin{figure*}[t]
\centering
\includegraphics[width=\textwidth]{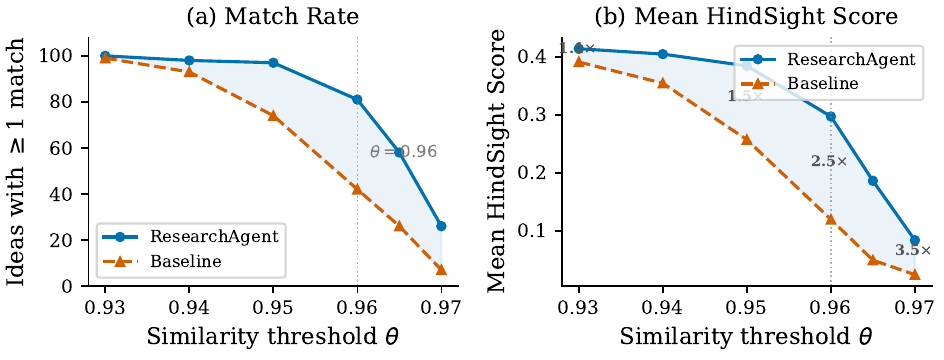}
\caption{Threshold sensitivity. \textbf{(a)}~At lenient thresholds ($\theta{\leq}0.93$) nearly all ideas match, reducing discriminative power. \textbf{(b)}~The ratio of RA to BL mean \hs{} scores grows monotonically from 1.1$\times$ to 3.8$\times$ as $\theta$ increases, confirming that the advantage is robust and amplified at stricter thresholds. Dotted lines mark $\theta{=}0.96$.}
\label{fig:threshold}
\end{figure*}

The similarity threshold $\theta$ is a key parameter.
Figure~\ref{fig:threshold} shows that the retrieval-augmented system's advantage is robust across the full range tested.

At lenient thresholds ($\theta{\leq}0.93$), almost all ideas match, collapsing the distinction.
As $\theta$ increases, the gap widens: the RA/BL ratio of mean scores grows from 1.1$\times$ at $\theta{=}0.93$ to 3.8$\times$ at $\theta{=}0.965$.
We select $\theta{=}0.96$ as the operating point because it provides strong discrimination (81\% vs.\ 42\% match rate) while retaining enough matches for statistical analysis.

\section{Case Study}
\label{sec:casestudy}

\begin{figure}[t]
\centering
\includegraphics[width=\columnwidth]{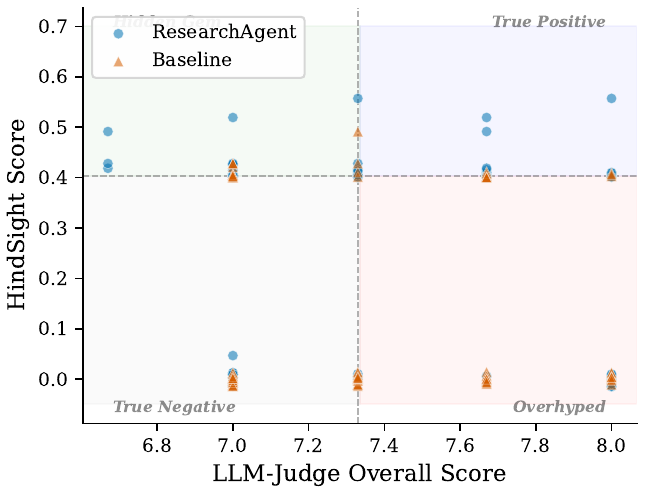}
\caption{Each idea plotted by LLM-Judge Overall ($x$) and \hs{} score ($y$). Dashed lines mark the medians used for quadrant classification. Retrieval-augmented ideas (blue) concentrate in the upper quadrants; baseline ideas (orange) cluster at $y{=}0$.}
\label{fig:scatter}
\end{figure}

To understand \emph{why} \hs{} and LLM-as-Judge disagree, we classify each idea into one of four quadrants based on whether its \hs{} score and LLM-Judge Overall score exceed their respective medians (Figure~\ref{fig:scatter}, Table~\ref{tab:casestudy}).

\begin{table}[t]
\centering
\small
\begin{tabular}{lcccc}
\toprule
& \multicolumn{2}{c}{\textbf{RA}} & \multicolumn{2}{c}{\textbf{BL}} \\
\cmidrule(lr){2-3} \cmidrule(lr){4-5}
\textbf{Category} & $n$ & \% & $n$ & \% \\
\midrule
True Positive & 17 & 17 & 6 & 6 \\
Hidden Gem & 23 & 23 & 13 & 13 \\
Overhyped & 26 & 26 & 34 & 34 \\
True Negative & 34 & 34 & 47 & 47 \\
\bottomrule
\end{tabular}
\caption{Quadrant distribution. \emph{True Positive}: high on both. \emph{Hidden Gem}: high \hs{}, low Judge (LLM underrates). \emph{Overhyped}: low \hs{}, high Judge (LLM overrates). \emph{True Negative}: low on both.}
\label{tab:casestudy}
\end{table}

\paragraph{True Positives (RA: 17, BL: 6).}
The retrieval-augmented system produces nearly 3$\times$ more true positives.
These ideas are technically concrete and well-grounded.
For example, an RA idea proposing a \emph{multimodal adapter framework for controllable text-to-image diffusion} matched IP-Adapter (1,397 citations) with similarity 0.977 and received a Judge score of 7.67---validated by both metrics.

\paragraph{Hidden Gems (RA: 23, BL: 13).}
23\% of RA ideas have high \hs{} but below-median LLM scores.
These tend to be technically specific ideas that lack the ``exciting narrative'' LLMs reward.
An idea about \emph{optimizing latent diffusion via cascaded architectures and knowledge distillation} scored only 7.33 from the judge but matched a real paper on self-cascade diffusion models (53 citations).
The LLM penalized it for perceived incrementality, yet it anticipated a concrete research direction.

\paragraph{Overhyped (RA: 26, BL: 34).}
Ideas that score well subjectively but zero on \hs{}.
These are characteristically \emph{grand but vague}: proposals for ``holistic frameworks'' or ``unified approaches'' that sound ambitious but are too abstract to match any specific paper.
A baseline idea proposing a \emph{game-theoretic framework for alignment in multi-agent systems} received 8.0/10 but matched nothing---too broad to constitute a publishable contribution.
The baseline has more overhyped ideas (34\% vs.\ 26\%), confirming that without retrieval grounding, the model produces more speculative ideas that score well subjectively.

\paragraph{True Negatives (RA: 34, BL: 47).}
Weak ideas correctly identified by both methods.
The baseline's larger share (47\% vs.\ 34\%) reflects that ideas without literature context are more often weak across both measures.

\subsection{Interpretation}

The case study reveals a systematic pattern: LLM judges reward \emph{novelty of framing} over \emph{anticipation of real impact}.
Ambitious, broadly-stated ideas score well but rarely correspond to specific research that gets published.
Meanwhile, technically grounded ideas---often enabled by literature retrieval---may appear incremental to an LLM but prove forward-looking.
This explains the negative $\rho$ with novelty: the most ``novel'' ideas (by LLM standards) tend to be the most speculative.

\section{Discussion}
\label{sec:discussion}

\paragraph{Complementary evaluation.}
Our results do not render LLM-as-Judge useless---it captures dimensions (clarity, ambition, coherence) that \hs{} cannot.
Rather, the two approaches are complementary: LLM-as-Judge for fast screening, \hs{} for objective validation when ground truth is available.
We recommend future work report both.

\paragraph{Implications for idea generation.}
If LLM judges consistently fail to distinguish retrieval-augmented from vanilla generation, then \emph{optimizing idea generation systems against LLM judge scores may be misguided}---it could push systems toward producing impressive-sounding but ultimately vacuous ideas.
\hs{} provides an alternative optimization target grounded in real impact.

\paragraph{Conformity bias.}
\hs{} rewards matching published work, potentially undervaluing truly revolutionary ideas that open entirely new directions.
We view this as an acceptable trade-off: the framework measures \emph{anticipation of real research trends}, which is meaningful even if incomplete.

\paragraph{False negatives.}
A zero \hs{} score does not mean an idea is bad---it may simply not have been pursued within the 30-month window, or the relevant papers may lie outside our 10-topic pool.

\paragraph{Embedding limitations.}
SPECTER2 captures semantic similarity at the topic level but may miss structural or methodological novelty expressed in different terminology.
Task-specific adapter heads \citep{singh2023scirepeval} or cross-encoder reranking could improve matching precision.

\section{Conclusion}

We introduced \hs{}, a time-split evaluation framework that measures research idea quality against real future publications.
Our experiments expose a fundamental disconnect: LLM judges see no difference between retrieval-augmented and vanilla idea generation, while \hs{} reveals a 2.5$\times$ gap.
The negative correlation between \hs{} and LLM-judged novelty suggests that language models reward ``novel-sounding'' framing over genuinely anticipatory thinking.
As AI idea generation systems grow more capable, grounding evaluation in real-world outcomes---rather than subjective impressions---will be essential.

\section*{Limitations}

\begin{itemize}[nosep,leftmargin=*]
  \item \textbf{Scale.}  We evaluate 200 ideas across 10 topics. Larger experiments with more systems would strengthen the findings.
  \item \textbf{Single embedding model.}  SPECTER2 base without adapter heads may miss nuanced matches. Cross-encoder reranking or ensembles could improve precision.
  \item \textbf{Ground truth completeness.}  Our pool of 27,589 papers does not cover all AI/ML research. Ideas matching uncovered papers receive false-zero scores.
  \item \textbf{Threshold sensitivity.}  While $\theta{=}0.96$ was calibrated empirically and results are robust (\S\ref{sec:threshold}), absolute scores depend on this choice.
  \item \textbf{Conformity bias.}  \hs{} rewards proximity to published work, potentially undervaluing genuinely novel ideas not yet explored.
  \item \textbf{Single judge model.}  We use Qwen3-32B as the sole judge; other models may yield different patterns.
\end{itemize}

\bibliography{custom}

\appendix

\section{Research Topics}
\label{sec:topics}

The 10 topics used in our experiments:
(1)~Alignment \& Safety,
(2)~Chain-of-Thought Reasoning,
(3)~Diffusion Models,
(4)~Efficient Inference,
(5)~Hallucination Mitigation,
(6)~In-Context Learning,
(7)~Instruction Tuning \& RLHF,
(8)~LLM Agents,
(9)~Multimodal LLMs,
(10)~Retrieval-Augmented Generation.
Each topic was selected to have significant pre-$T$ literature ($\geq$50 papers) and substantial post-$T$ developments ($\geq$100 papers in the ground truth pool).

\section{LLM-as-Judge Prompt}
\label{sec:judgeprompt}

The system prompt instructs the judge to evaluate each idea on four dimensions (1--10):
\textbf{Novelty} (originality beyond incremental extensions),
\textbf{Feasibility} (practicality with current tools and data),
\textbf{Expected Impact} (significance if successful), and
\textbf{Overall} quality.
The judge responds in JSON with integer scores and a brief rationale.
Each idea receives 3 independent evaluations (temperature 0.7), with scores averaged.

\section{Impact Score Details}
\label{sec:impact}

Impact scores in the ground truth pool range from 0 to 1, with a mean of 0.03 and median of 0.006 (heavily right-skewed due to power-law citation distributions).
Papers published at the 7 top venues ($v(p){=}1$) account for approximately 8\% of the pool.
The 0.6/0.4 weighting between citations and venue was chosen to balance quantitative impact with peer-review quality signal.

\section{Correlation Heatmap}
\label{sec:heatmap}

\begin{figure}[h]
\centering
\includegraphics[width=\columnwidth]{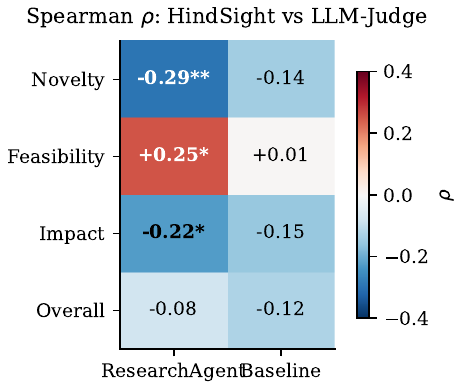}
\caption{Spearman $\rho$ between \hs{} and LLM-Judge dimensions. Stars denote significance: {*}~$p{<}0.05$, {**}~$p{<}0.01$, {***}~$p{<}0.001$.}
\label{fig:correlation}
\end{figure}

Figure~\ref{fig:correlation} visualizes the Spearman correlations from Table~\ref{tab:correlation}.
The negative correlations for ResearchAgent (blue cells) and the absence of significant correlations for the baseline (near-white cells) are clearly visible.

\end{document}